\documentclass[letterpaper, 10pt, twocolumn]{article}
\usepackage{clean}
\shorttitle{Grasp stability and slip}

\usepackage{graphicx}
\usepackage{svg}
\usepackage{subfig}
\graphicspath{{figs/}}
\usepackage{nicematrix}
\usepackage{array}
\usepackage{multirow}
\usepackage{caption}

\usepackage{amsmath,amssymb,amsfonts}
\usepackage{algorithm2e}
\usepackage{gensymb}
\usepackage{xcolor}
\usepackage[colorlinks=true, linkcolor=black, 
            urlcolor=cyan, filecolor=black,
            citecolor=black]{hyperref}
\usepackage{url}
\usepackage{amsmath,bm}
\usepackage{multicol, multirow, makecell} %
\usepackage{comment}
\usepackage{algorithm2e}

\usepackage{cite}

    \title{Soft finger rotational stability for precision grasps}
\author{Hun Jang$^{1*}$, Valentyn Petrichenko$^{1*}$, Joonbum Bae$^2$, Kevin Haninger$^1$ 
\thanks{$^{1}$Department of Automation, Fraunhofer IPK, Berlin, Germany
        {\tt\small firstname.lastname@ipk.fraunhofer.de}}%
\thanks{$^{2}$ Department of Mechanical Engineering, UNIST, Ulsan, Korea {\tt\small jbbae@unist.ac.kr}}
\thanks{$^{*}$ denotes equal contribution. This research was supported by the MOTIE (Ministry of Trade, Industry, and Energy) in Korea, under the Fostering Global Talents for Innovative Growth Program related to Robotics (P0017311) supervised by the Korea Institute for Advancement of Technology (KIAT)}
}
\begin{document}
\maketitle 
\begin{abstract}
Soft robotic fingers can safely grasp fragile or variable form objects, but their force capacity is limited, especially with less contact area: precision grasps and when objects are smaller or not spherical. Current research is improving force capacity through mechanical design by increasing contact area or stiffness, typically without models which explain soft finger force limitations. To address this, this paper considers two types of soft grip failure, slip and dynamic rotational stability. For slip, the validity of a Coulomb model investigated, identifying the effect of contact area, pressure, and relative pose. For rotational stability, bulk linear stiffness of the fingers is used to develop conditions for dynamic stability and identify when rotation leads to slip. Together, these models suggest contact area improves force capacity by increasing transverse stiffness and normal force. The models are validated on pneumatic fingers, both custom PneuNets-based and commercially available. The models are used to find grip parameters which increase force capacity without failure.
\end{abstract}

\section{Introduction}
Soft fingers can grasp sensitive objects and handle geometry variation, especially with spherical or cylindrical objects where an enclosing grasp can be used \cite{ozawa2017,hernandez2023}. An enclosing or power grasp gives a large contact area between finger an object, and is well-suited to pick-and-place tasks in unobstructed environments \cite{piazza2019}. 

However, soft fingers typically have reduced force capacity compared with rigid fingers \cite{li2021a, liu2021}, limiting payload \cite{azami2023} and contact-rich tasks \cite{billard2019}. This limitation is significant when contact area is reduced: when smaller or flat objects are grasped, or when a precision grasp, such as fingertip or pinch grasp, is needed, e.g. to manipulate dexterously in cluttered environments \cite{teeple2020multi}.  Although three or more fingers can be used to improve grasp, objects which are smaller, not round, or in cluttered environments often require a parallel grasp with two fingers. 

Ongoing work has explored increasing grasp strength through mechanical design: using stiffening elements \cite{park2020}, jamming techniques \cite{singh2023}, incompressible actuating fluids \cite{azami2023}, additional kinematic structure \cite{zhou2017}, and structured compliance \cite{hartisch2023}. While mechanical design is important, the force capacity is also affected by grip parameters such as degree of actuation, robot grip pose, and finger distance. 

\begin{figure}
    \centering
    \captionsetup{justification=centering}
    \subfloat[Rotational instability on a range of fingers and objects]{\includegraphics[width=\columnwidth,height=3cm]{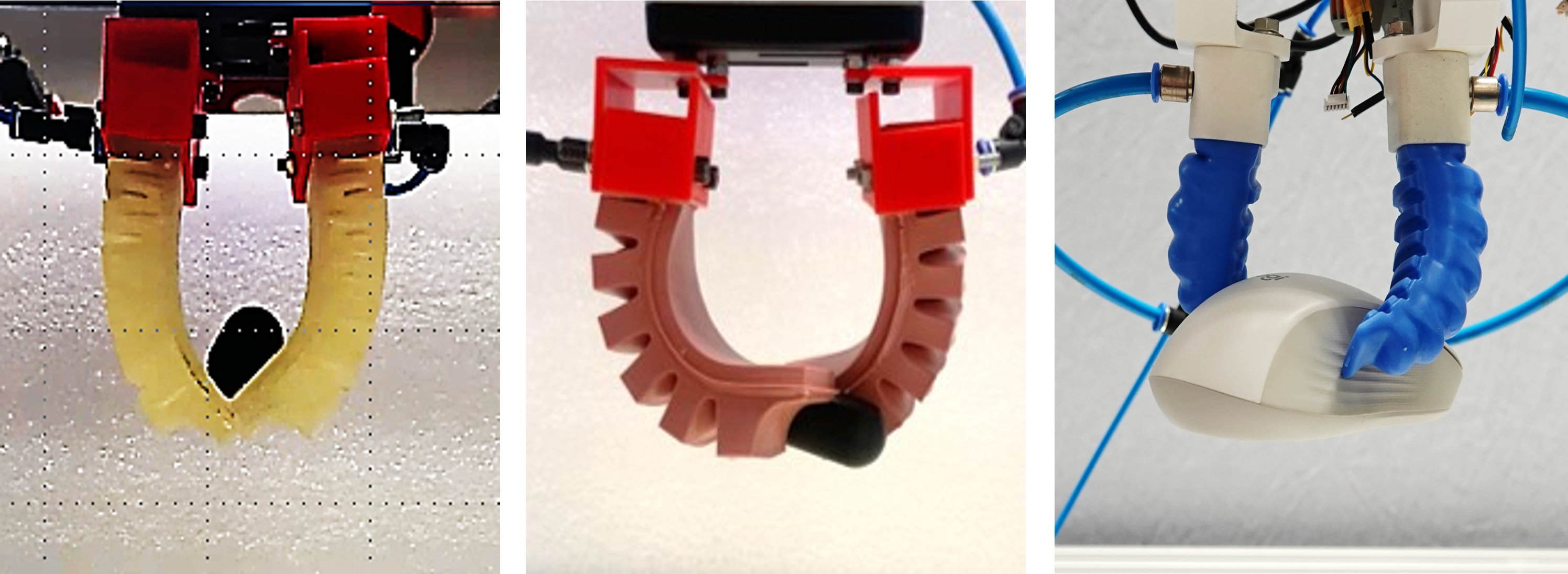}} \label{fig:grip1} 
    \subfloat[Transition as pressure increases]{\includegraphics[width=\columnwidth,height=3cm]{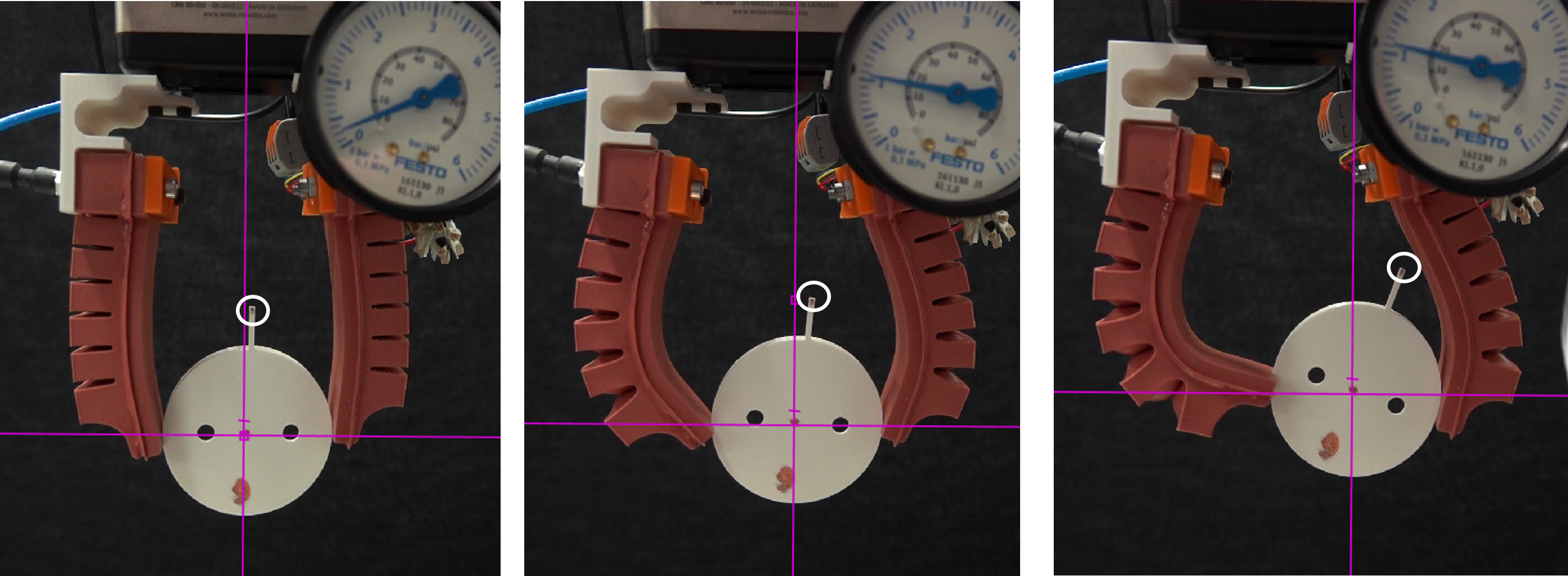}} \label{fig:grip3}
\caption{Rotational grasping instability of soft robotic fingers}
\label{fig:dynamic_stability}
\end{figure}

Grasping instability can be seen on a range of fingers and objects in Fig.~\ref{fig:dynamic_stability}(a). When pressure increases, as seen in Fig. ~\ref{fig:dynamic_stability}(b), the object begins to rotate, sometimes leading to slip at the contact with the fingers.  We observe that this rotational instability limits the ability to increase force capacity through increasing pressure, and that it is affected by grasp parameters. Due to the complexity of soft fingers, tuning these grip parameters often relies on the experience of the users or computational simulation \cite{graule2021somo}.

This rotational instability is influenced by both grip stability and slip. Grip stability has a wide range of definitions \cite{roa2015, prattichizzo2008}. Planar grip stability has been studied for force-closure \cite{trinkle1994}, form-closure \cite{bicchi2000}, and stiff-but-underactuated fingers \cite{birglen2007}. However, these analysis assume the contact is a point and infinitely stiff, making them unrealistic for soft fingers. Some approaches do model finger compliance \cite{prattichizzo2008} for tendon-driven fingers \cite{haas-heger2018} or spherical fingertips \cite{dong2019}. However, soft fingers can have rotational instability \cite{teeple2020multi}, as seen in Figure~\ref{fig:dynamic_stability} and the attached video, which requires modelling bulk motion of the fingers.

Slip is typically modelled with Coulomb friction in grasping \cite{roa2015, liu2023}. However, a Coulomb friction model is a point-contact model, raising questions about its applicability when a larger contact area in soft fingers. Friction models for spherical soft fingertips have been proposed \cite{fakhari2016, fakhari2019, dong2019}, but not validated on general soft fingers.

Towards increasing the force capacity of soft fingers, this paper proposes a model for rotational stability and validates the bulk Coulomb model. Compared with friction modelling in soft fingers \cite{liu2023, fakhari2019}, we examine the effect of contact area, pressure, and offset. This suggests increasing normal force is the primary way to increase force capacity. However, increasing normal force can lead to dynamic instability. Rotational instability is modelled considering linear finger stiffness. Compared with classical grip stability \cite{roa2015, trinkle1994, bicchi2000, birglen2007, haas-heger2018}, this is the rotational stability. These models are validated experimentally on a range of objects, grip conditions, and soft fingers (PneuNet \cite{polygerinos2015a} and commercial fingers). We then use the model to find grip parameters which improve force capacity.

In summary, we make following contributions: 1. an analytical model of rotational stability in soft finger’s precision grasping relating rotation angle to bulk stiffness; 2. empirically finding the relationship between grasp parameters and the model variables, validating coulomb friction and finding effects on bulk stiffness; 3. validating the model by comparing the force/angle between the model prediction and actual results and introduce a simple case using the model for optimizing the grasping parameters.

\section{Dynamic grip stability \label{sec:dyn_stability}}
\begin{figure}[h]
    \centering
    \includegraphics[width=0.8\columnwidth]{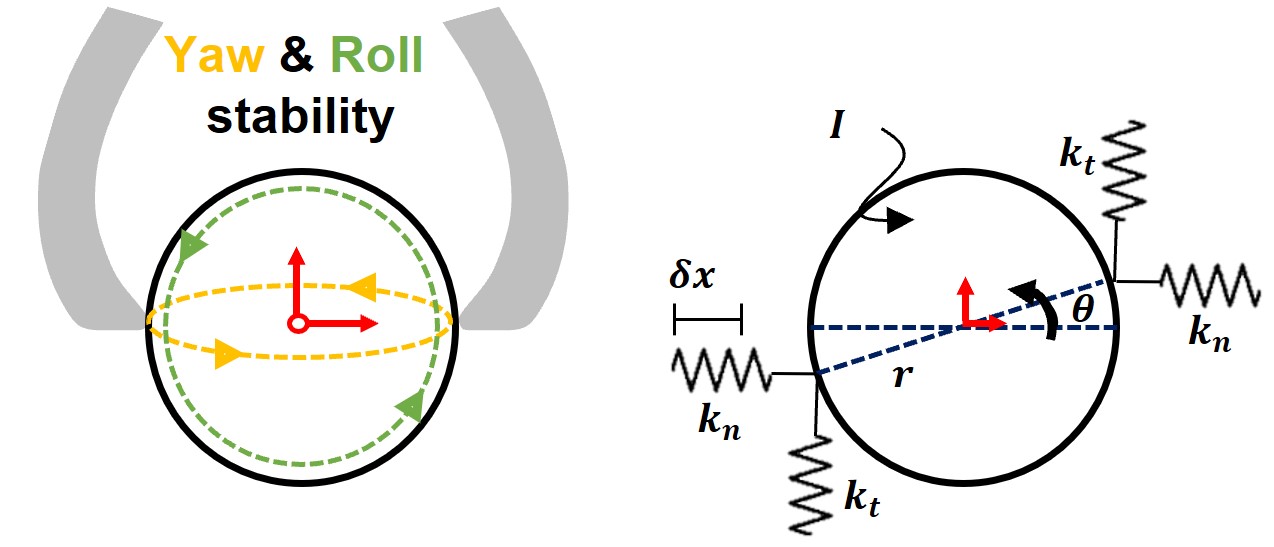}
    \caption{Stiffness measurement of a planar soft finger, where the $\cdot_n$ and $\cdot_t$ denote normal and transverse, which change spatial direction for instability about the $z$ or $x$ axis.}
    \label{fig:grip_spring_model}
    \vspace{-10pt}
\end{figure}

We consider a planar precision grasp as seen in Figure \ref{fig:grip_spring_model}, which can be the vertical(roll) or horizontal(yaw) planes. We define rotational stability of soft fingers as the convergence of $\theta \rightarrow 0$ over time, and derive stability conditions about the equilibrium point $\theta=0$ using linear stability of the dynamic system.

The soft fingers are modelled as a 2D linear stiffness which couples the object to the robot flange. This stiffness is parameterized by $(k_n, k_t)$ in the normal and transverse directions. We assume $k_n$ and $k_t$ are equal between the fingers, ignore any torsional stiffness at the contact area due to a smaller contact area(precision grasping), and assume that the direction of the normal and transverse force stays constant during rotation. We assume both fingers contact at a radius $r$ from the center of mass of the object. Grip force of finger in the normal direction is modeled by a relative spring displacement $\delta_n$ and preload in the transverse direction is assumed to be $0$.

\subsection{Rotational dynamics}
We model the pure rotation of the object seen in Figure \ref{fig:grip_spring_model}. The dynamics can be determined based on the energies of the system with Lagrangian mechanics. The kinetic energy $T$ in the case of a pure rotation is given by $T=\frac{1}{2}I\dot{\theta}^2$, where $I$ is the rotational inertia and $\dot{\theta}=d\theta/dt$. The potential energy of the system $V$ is derived as
\begin{align}
     V=k_n(\delta_n -r(1-\cos{\theta}))^2+k_t(r \sin{\theta})^2, 
      \label{eq:Potential}
\end{align}
where $r$ is object radius, $k_n$ and $k_t$ the stiffness in respective directions, and $\delta_n$ the preload displacement. We then find equations of motion of
\begin{equation}
    \begin{split}
        I \ddot{\theta}= & 2k_nr \sin{\theta} \delta_n-2k_n r^2(1-\cos{\theta})\sin{\theta}\\
                        & -2k_t r^2 \sin{\theta} \cos{\theta}.
    \end{split}
    \label{dyn_eq}
\end{equation} 

\subsection{Linearized dynamics and stability}
With a state vector $x=[\theta, \dot{\theta}]^T$, the convergence of $x\rightarrow 0$ can be considered from the control theory. When the system is dynamically stable about an equilibrium point (here, $x=0$), it converges to that state over time from starting conditions near that state. When the system is dynamically unstable, it will diverge.  

While the dynamics \eqref{dyn_eq} are nonlinear, the local stability can be analyzed from linearizing about $x=0$, yielding a system matrix $A$ of
\begin{align}
    A=\begin{bmatrix}
    0 & 1 \\
    2rI^{-1}(k_n \delta_n - k_tr) & 0
\end{bmatrix},
\end{align}
where $\dot{x} \approx Ax$. The stability of the linearized system is given by the eigenvalues of the $A$ matrix, which can be found as
\begin{equation}
    \begin{split}
    \lambda_{1/2} & = \pm \sqrt{2rI^{-1}(k_n \delta_n - k_tr)} \\ 
                   & =\begin{cases}
                           \pm \sqrt{2rI^{-1}(k_n \delta_n - k_tr)}, \text{ for }  k_n \delta_n > k_tr 
                           \\ \pm i \sqrt{2rI^{-1}(k_tr - k_n \delta_n)}, \text{ for }  k_n \delta_n < k_tr
                      \end{cases}.
    \end{split}
\end{equation}
If the normal force at $\theta=0$, denoted preload force $f_p = k_n \delta_n$, is greater than the product of transverse stiffness $k_t$ and object radius $r$, a pole will be in the right half-plane, indicating instability. This gives a dynamic stability condition about the equilibrium point $\theta=0\degree$ of
\begin{align}
    \underbrace{k_n \delta_n}_{f_p} < k_tr,
    \label{eq:condition}
\end{align}
and we denote the force when this condition is violated as $f_p^i = k_t r$.

\subsection{Rest position}
As the preload force increases, at $f_p \geq f_p^i$ rotation away from $\theta=0$ begins. However, the finger may converge to a rest angle $\theta \neq 0$, which we denote $\theta_r$. We find this rest angle from \eqref{dyn_eq} by setting $\ddot{\theta}=0$ and solving for $\theta$. This system has solutions at $\theta_r=n\pi$ for $n\in\mathbb{Z}$, and additional solutions when 
\begin{equation}
    \left\vert \frac{k_n \delta_n - k_n r}{k_t r-k_n r} \right\vert < 1.  \label{eq:rest_condition}
\end{equation} 

\begin{figure}[t]
    \centering
    \includegraphics[width=0.75\columnwidth]{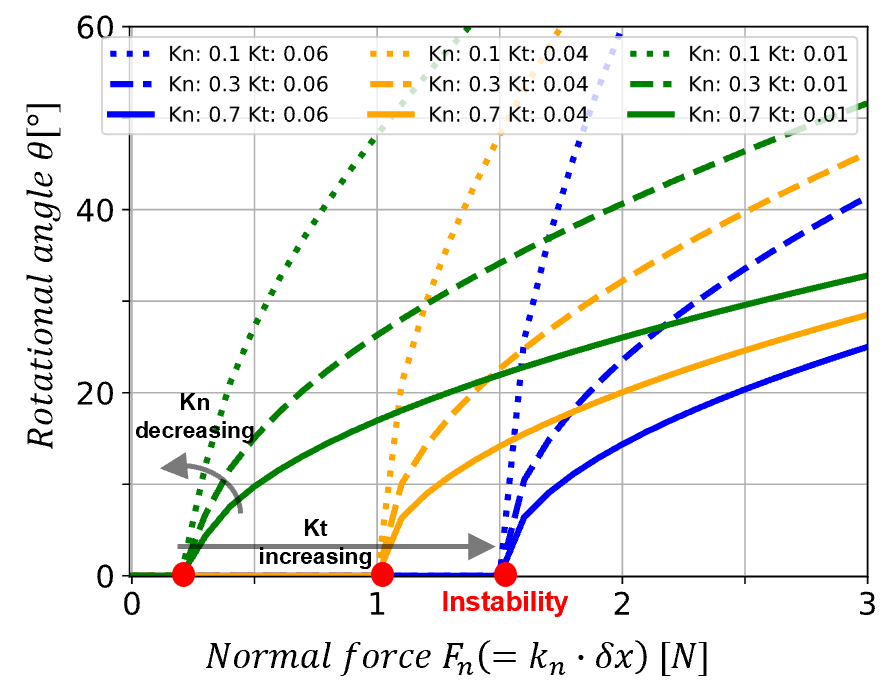}
    \caption{Rest angle $\theta_f$ progression as  $k_n$ and $k_t$ vary.} 
    \label{fig:stiffness_comp}
\end{figure}

The progression of $\theta_r$ as $f_p$ increases is shown for selected system parameters in the Fig. \ref{fig:stiffness_comp}. It can be seen that the object begins to rotate when $f_n$ exceeds the limit in \eqref{eq:condition}. It should be noted that if the condition \eqref{eq:rest_condition} is not satisfied, the progression of the rest position angle cannot be calculated and only the force $f^i_p$ can be determined. 

The comparison of these results for different $k_n$ and $k_t$ values can be seen in Fig. \ref{fig:stiffness_comp}. As indicated by \eqref{eq:condition}, the initiation of rotation depends on $k_t$, increasing as $k_t$ increases. However, the slope after rotation is initiated decreases as $k_n$ increases.

\subsection{Rotation leading to slip}
As the object rotates, the normal and transverse forces change, possibly leading to slip which often results in small objects flying from the gripper. Fig. \ref{fig:grip_spring_model} shows the normal force $f_n$ and static friction force $f_t$, which from the nonlinear model can be defined as follows: 
\begin{eqnarray}
    f_n&=&k_n(\delta_n -r(1-\cos{\theta})) \\ &=& f_{p}-k_n r(1-\cos{\theta}) \\
    f_t&=&k_tr \sin{\theta},
\end{eqnarray}
where $f_p=k_n \delta_n$ is the preload force on the finger.
These equations are connected by the Coulomb friction condition $|f_t| \leq f_n\mu$. With the assumption that the direction of the forces $f_n$ and $f_t$ doesn't change with rotation, the limiting case for the angle from which the slip begins can be determined by solving $f_t = \mu f_n$, which gives \cite{asinxbcosx}
\begin{equation}
    \theta_f=2(\arctan(\frac{a\pm\sqrt{a^2+b^2-c^2}}{b+c})+n\pi), n \in \mathbb{Z},
    \label{eq:friction_angle}
\end{equation}
where $a=k_tr, b=-\mu k_nr$ and $c=\mu f_p-\mu k_nr$. The angle $\theta_f$ indicates the rotation at which slip occurs. It follows that slip does not occur if the rest angle  remains smaller than the critical angle $\theta_r<\theta_f$. %

\section{Friction modelling}
\label{sec:friction_modelling}
This section introduces the friction model, validating the bulk Coulomb model for soft fingers. Coulomb friction models slip as a violation of the condition $|f_t| < \mu f_n$, where $\mu\geq 0$ is the Coulomb friction coefficient, $f_n$ normal force and $f_t$ tangential force. These models are popular in soft robotics \cite{chavan-dafle2019, park2020, hogan2020}, and can also be made stochastic \cite{liu2023}. 

\begin{figure}[t]
    \centering
    \subfloat[Soft fingers, (a) are PneuNets and (b) are from SoftGripping and test object on the F/T sensor]{\includegraphics[width=0.75\columnwidth]{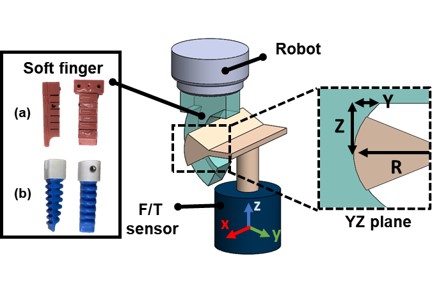}} \\
    \subfloat[Friction measurement motion and representative results]{\includegraphics[width=0.75\columnwidth]{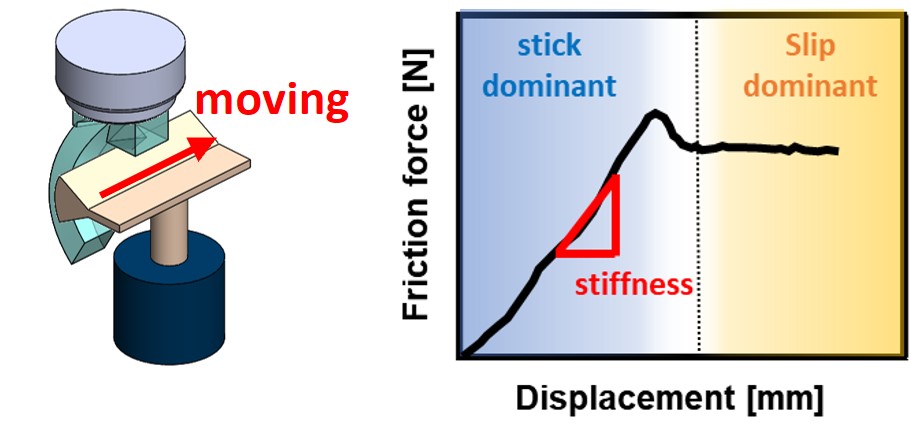}}
    \caption{Friction measurement experimental setup}
    \vspace{-10pt}
    \label{fig:experiment_setup}
\end{figure}

However, the Coulomb model is a point model, and it is unclear how it applies to larger contact areas. One reason for this is that the pressure distribution over the contact area not uniform, which can result in local slip, where local relative motion occurs before bulk slip \cite{fakhari2016}. Computational methods to estimate contact pressure have been proposed \cite{masterjohn2022}, but not yet extended with frictional models. To investigate the accuracy of the bulk Coulomb friction over soft grasp parameters (contact area, offset, pressure, material), we develop a test environment as seen in Figure \ref{fig:experiment_setup}.

\begin{table}[h!]
    \centering
    \begin{tabular}{|r|l|}
        \hline
        \textbf{Parameter name} & \textbf{Value} \\
        \hline\hline
        \textbf{Pressure [\(bar\)]} & 0.4, 0.8, 1.2 \\
        \hline
        \textbf{Contact area [\(cm^2\)]} & 2.1(S), 4.2(M), 8.4(L) \\
        \hline
        \textbf{Horizontal offset [\(mm\)]} & 0, 10, 20 \\
        \hline
        \textbf{Vertical offset [\(mm\)]} & -20, 0, 20 \\
        \hline
        \textbf{Contact surface material friction} & Ordinary, High \\
        \hline
    \end{tabular}
        \caption{Parameters of the friction experiments\label{tab:Parameter_experiment}}
\end{table}

\subsection{Coulomb model validation}
The experimental setup seen in Figure \ref{fig:experiment_setup}, where cylinder segments of various arc lengths are 3D printed from Tough PLA with a radius $30$ mm.  In this way, the contact area is controlled by the test object, provided the finger is in contact with the full arc of the object, which is verified visually per experiment.  The test objects are mounted onto a Force/Torque sensor (ME-Me\ss systeme MP11, load limit 500N, 20Nm, sampling rate 125Hz) to allow measurement of the total normal and transverse force.

The pneumatic fingers are made with Wacker M4601 Elastosil, and the kinematic structure is based on PneuNets \cite{polygerinos2015a}. They are actuated by controlled pressure valves (VEAB-L-26-D9-Q4-V1-1R1, Festo), where a desired pressure is applied. The fingers are mounted on a Universal Robots UR10, which is used to initiate contact and load the fingers. We then compare five different parameters: pressure, contact area, relative position between finger and contact object (horizontal and vertical offset) and contact surface material. The range of parameters tested is shown in Table \ref{tab:Parameter_experiment}. The high friction condition adds a rubber to the surface of the test object.

For each set of parameters, the following experiment process is followed: (i) the finger is actuated in free space, (ii) the robot moves the finger into contact, (iii) contact along the entire test object is visually verified, (iv) linear motion in the positive $y$ direction begins, with a velocity of $2$mm/s and to a distance of $30$mm, (v) the robot moves back to the initial pose. As the robot moves in the $y$ direction, the total transverse force is measured as $f_t=f_y$ from the force/torque sensor. The total normal force is calculated as $f_n = \sqrt{f_x^2+f_z^2}$, where $f_z$ is assumed to be from asymmetrical contact pressure, not friction. Experiments are conducted for each parameter combination three times, with the mean value plotted. 

Three phases of the friction forces can be identified, illustrated in Figure \ref{fig:experiment_setup}(b), right: the stick phase, where the fingertip is not moving while the robot moves with constant velocity, a transition region, and a constant friction phase as sliding occurs. In the stick phase, the slope of $f_t$ represents the soft finger's bulk stiffness in $y$, $k_y$. After the transition phase, the asymptote in the sliding phase is taken as the friction coefficient $\mu$. 

\begin{figure}[t]
    \centering
    \subfloat[P: pressure, P(High Fc): high $\mu$ surface, CA: contact area]{\includegraphics[width=0.49\columnwidth]{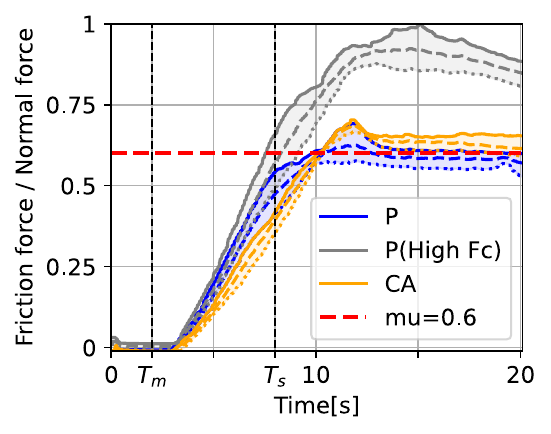} \label{fig:subim1}}  
    \subfloat[Horizontal offset and Vertical offset]{\includegraphics[width=0.49\columnwidth]{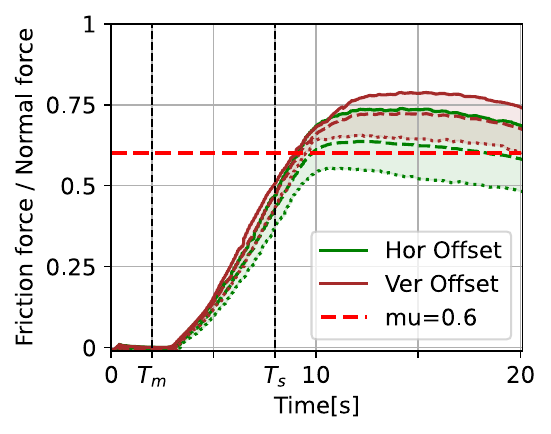} \label{fig:subim2}}
\caption{Variation in friction coefficient $f_t/f_n$, where the shaded region shows min/max values for specified grasp parameters. The reference value of $\mu=0.6$ is shown in red.}
\label{fig:image3}
\vspace{-15pt}
\end{figure}

The transverse force over normal force, $f_t/f_n$, can be seen in Figure \ref{fig:image3}, showing the min/max over each parameter in the shaded region. These results suggests that material choice dominates $\mu$, which is less affected by other grip parameters. The vertical and horizontal offsets increase variation in $\mu$ due to changing the pressure profile, but these effects are bounded within $\mu \in [0.49, 0.77]$. While vertical and horizontal offset affect the effective friction coefficient, the varation in $\mu$ is $\pm 23\%$, and total friction force $f_t$ can easily be dominated by changes in $f_n$. Consequently, the time interval ranging from $T_m$ to $T_s$ can be characterized as the stick-dominant phase, representing the measurable stiffness range.

\section{Experimental validation of stability models}
To validate the stability conditions in \ref{sec:dyn_stability}, the stiffness properties are measured, then the rotation of test objects as pressure increases is shown for a range of grasp conditions. This is used to evaluate the models in Section \ref{sec:dyn_stability}: when rotation away from initial orientation begins $f^i_p(k_n, k_t, r)$, the progression of rest angle $\theta_r$, and when the rotation leads to slip at $f^s_p(k_n, k_t, r)$. 

\subsection{Measuring Finger Stiffness}

\begin{figure}[t]
    \centering
    \subfloat[Red fingers, horizontal offset, where rotation about $x$ means $k_n=k_y$, $k_t=k_z$]{\includegraphics[trim={0.2cm 0 1cm 1cm}, clip, width=0.8\columnwidth]{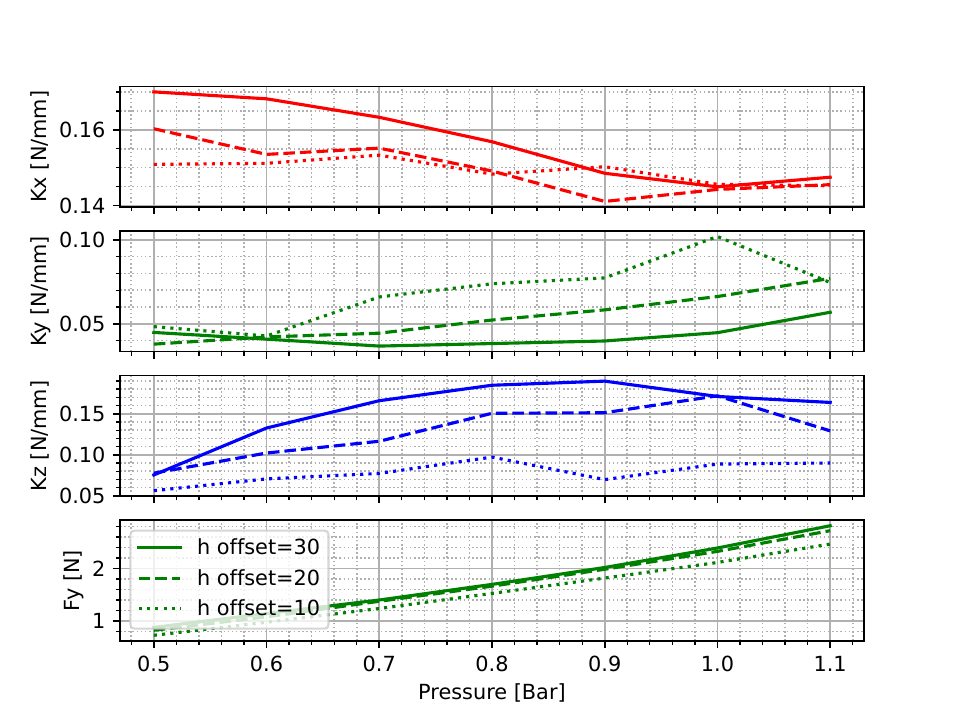} \label{fig:stiff_red}} \\
    \subfloat[Blue fingers, vertical offset, where rotation about $z$ means $k_n=k_y$, $k_t=k_x$]{\includegraphics[trim={0.2cm 0 1cm 1cm}, clip, width=0.8\columnwidth]{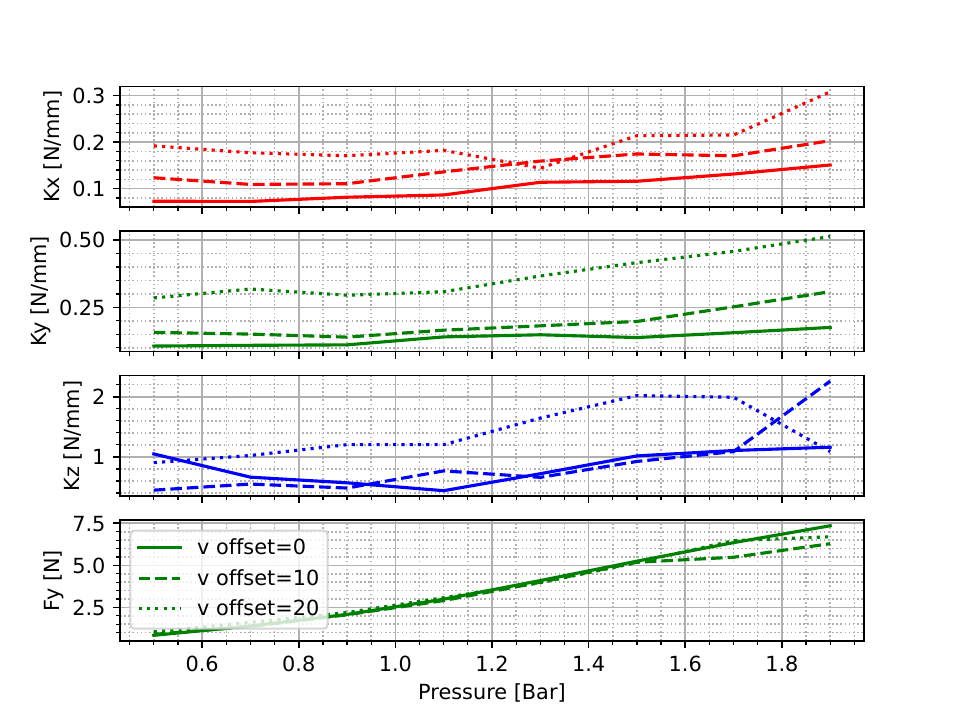} \label{fig:stiff_blue}}
    \caption{Finger stiffness and preload force over pressure and offset for the (a) red PneuNet fingers and (b) blue Soft Gripping fingers.\label{fig:stiff_measure}}
\vspace{-10pt}
\end{figure}

\begin{figure}[t]
    \centering
    \subfloat[Red fingers with $60$mm object and horizontal offset $10$mm (left), $20$mm (mid), $30$mm (right)]{
        \includegraphics[height=3.5cm]{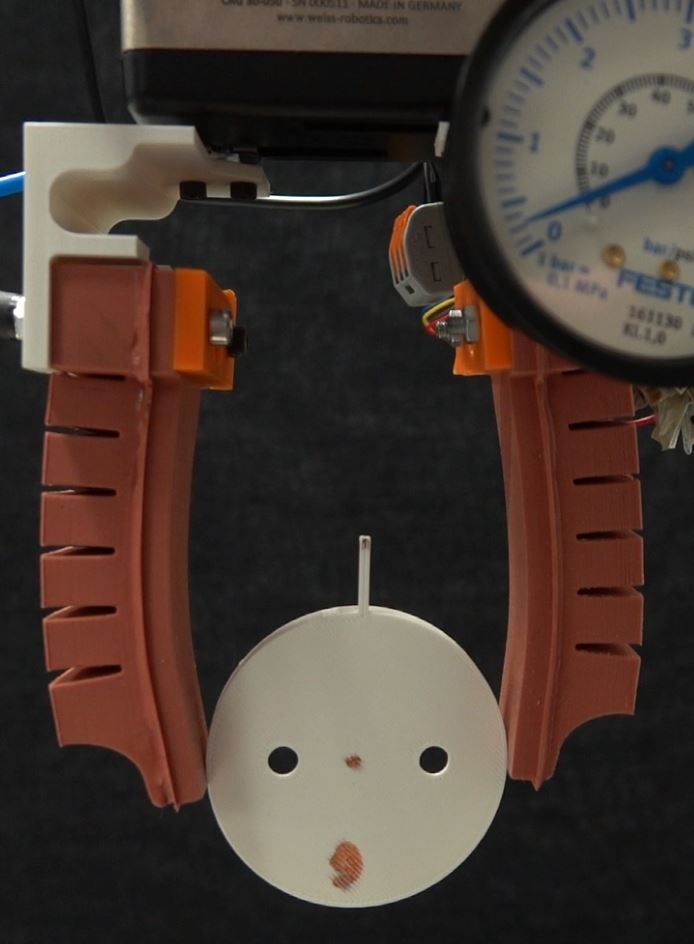}
        \includegraphics[height=3.5cm]{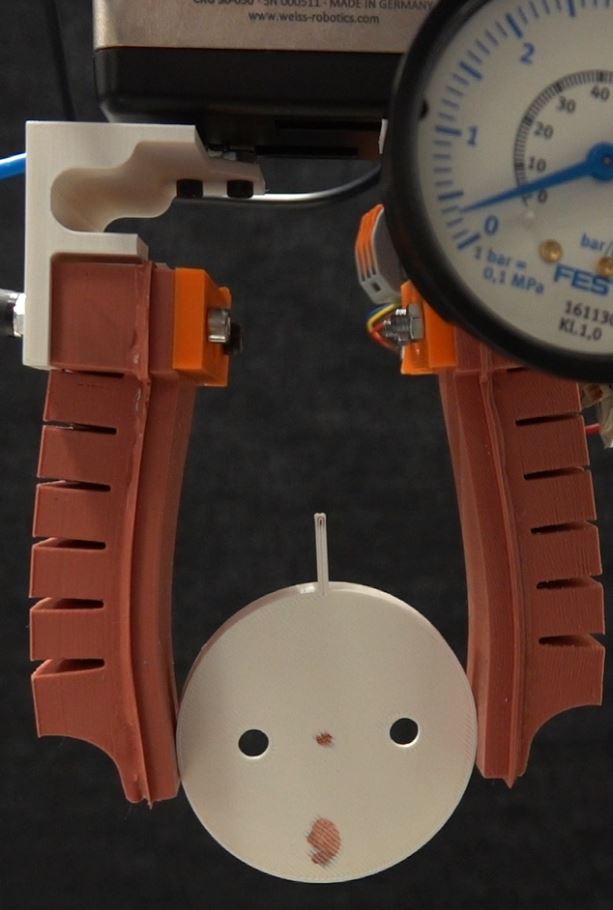}
        \includegraphics[height=3.5cm]{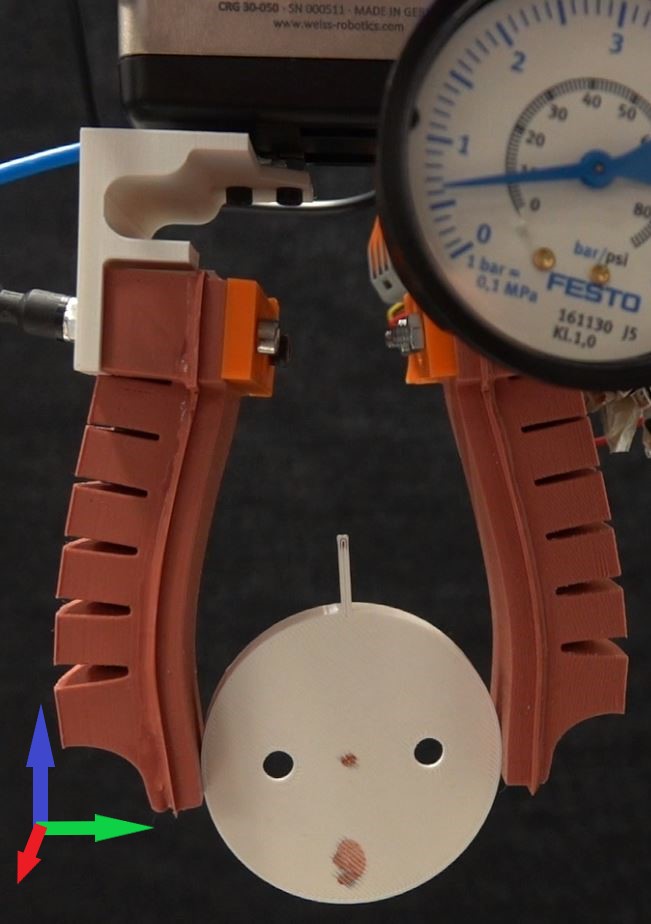} 
    } \\
    \subfloat[Blue finger vert offset $0$mm (left), $10$mm (mid), $20$mm (right)]{        
        \includegraphics[trim={46cm 8cm 30cm 10cm}, clip, height=3cm]{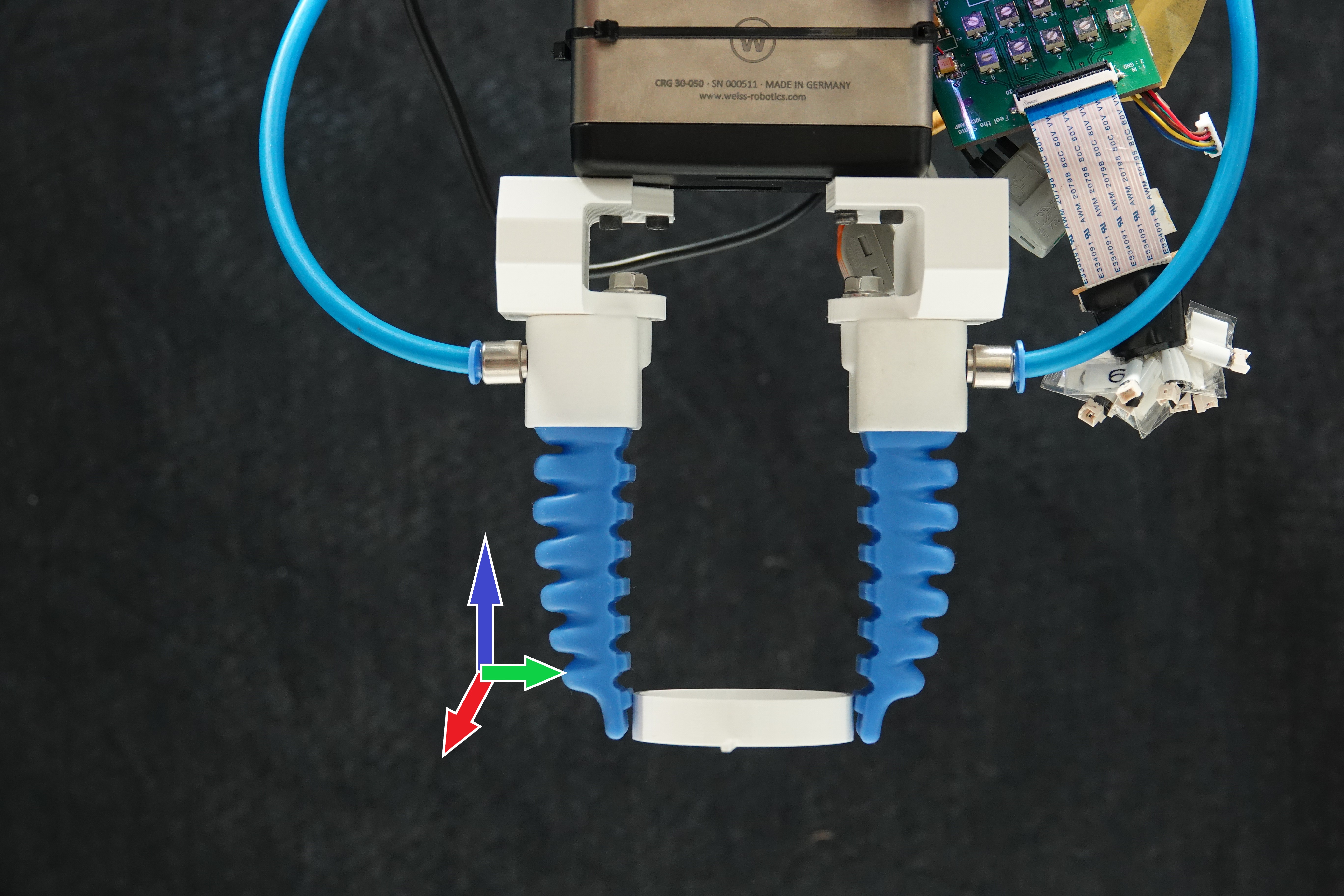}
        \includegraphics[trim={13cm 3cm 9cm 3cm}, clip, height=3cm]{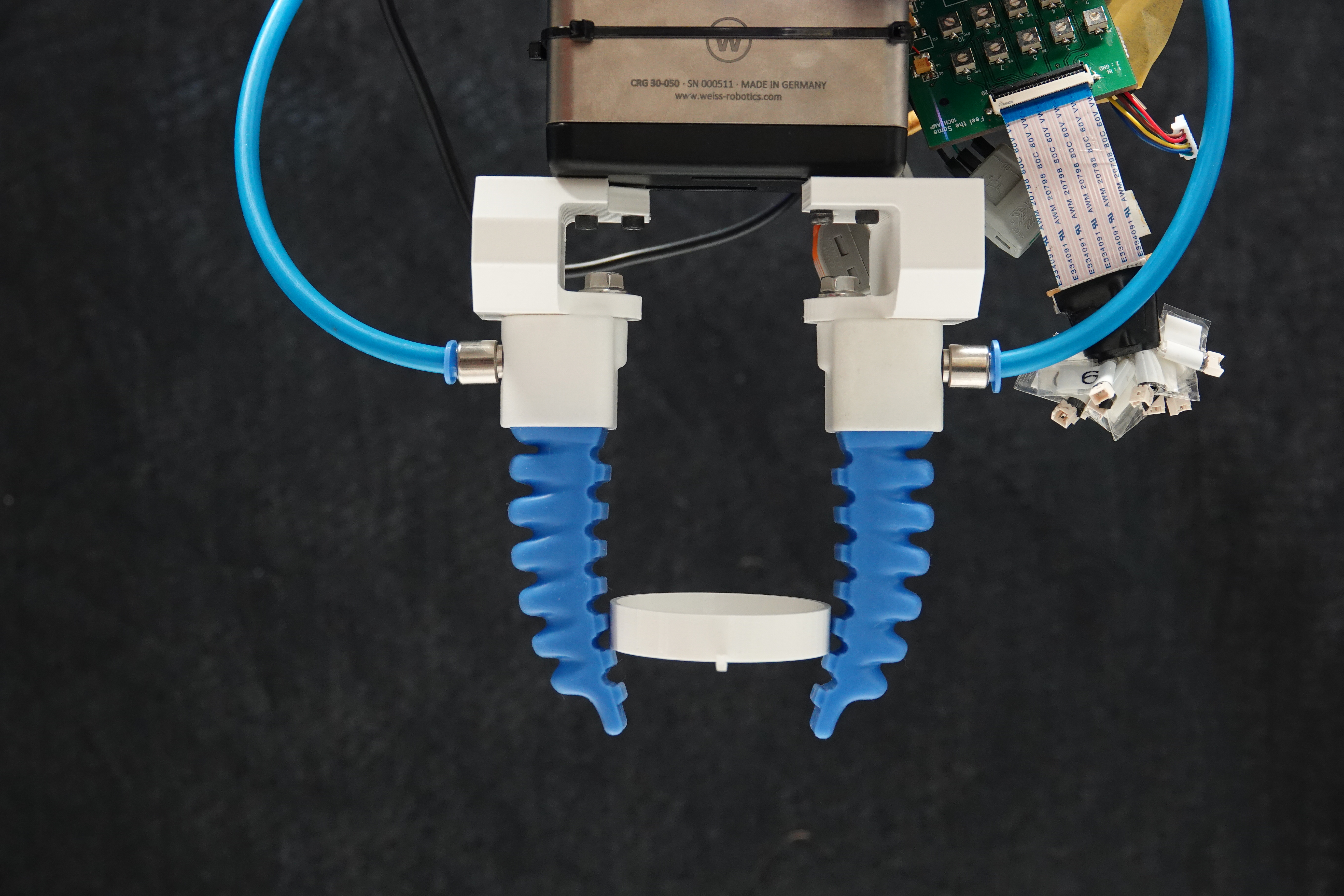}
        \includegraphics[trim={13cm 3cm 9cm 3cm}, clip, height=3cm]{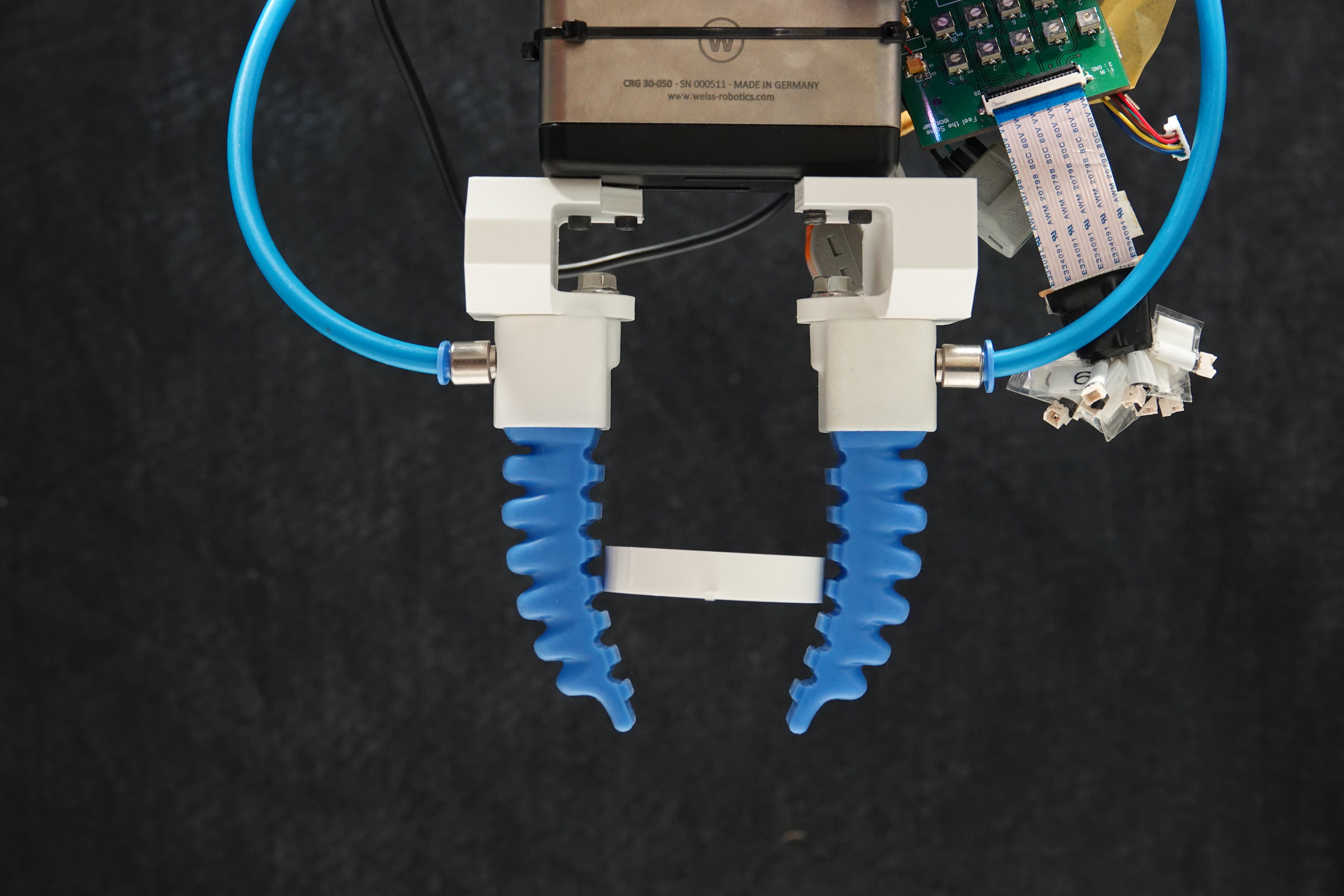} }
    \caption{Grasp conditions for the rotation angle experiments, coordinates shown with $x$ red, $y$ green and $z$ blue.  In addition to the $60$mm diameter object, $50$ and $70$ are used. \label{fig:instabil_conditions}}
\end{figure}

The stiffness of the finger is needed to validate the stability model. In order to measure this, an object is fixed on the force sensor and a single finger mounted on the robot, as seen in Figure \ref{fig:experiment_setup}. A range of vertical or horizontal offsets are applied by moving the robot / gripper, then, for each offset, a range of pressures is applied. When the pressure is applied, the robot makes a small relative movements in the three spatial directions, pauses, and the forces are measured. For each motion, e.g. from $x_0$ to $x_1$, the forces measured at each one as $f_0$ and $f_1$, and the corresponding linear stiffness can be found as $k=\frac{f_1-f_0}{x_1-x_0}$. The size of the deltas is $[2, 2, 0.5]$mm (in $x$, $y$ and $z$ related to force sensor coordinate system) for the blue SoftGripping fingers \cite{softgripper}, and $[2, 2, 1]$mm for the red PneuNet fingers \cite{polygerinos2015a} to avoid slip. These measurements are repeated at a range of pressures and robot pose offsets which correspond with the experiment conditions seen in Figure \ref{fig:instabil_conditions}.

The results can be seen in Figure \ref{fig:stiff_measure}(a) for the PneuNets (red) and Figure \ref{fig:stiff_measure}(b) for the SoftGripping fingers (blue), under horizontal and vertical offsets as seen in Figure \ref{fig:instabil_conditions}. A few general conclusions can be made: the offsets do not majorly change the relation between pressure and $f_y$. The stiffness in grip direction $k_y$ is affected by pressure. For the red PneuNets, the vertical stiffness $k_z$ increases as the offset increases, i.e. either a larger object is grasped or the parallel gripper is closed. However, $k_x$ is not affected by offset.  For the blue Soft Gripping fingers, the horizontal stiffness $k_x$ is affected by vertical offset, increasing as the object is grasped `deeper' in the fingers. $k_z$ is large, and thus instability about $x$ (where $k_t=k_z$) would not be expected.

\subsection{Rotational measurements}
To validate the rest angle that the object takes as grip force is increased, we use the test objects and grasp conditions seen in Figure \ref{fig:instabil_conditions}, and take a video as the pressure is increased by steps. The red dots on the indicating stick and center can be tracked in the video software Tracker\footnote{https://physlets.org/tracker/}, and are used to find the change in rotation angle $\theta$. To explore various experimental scenarios, we conducted experiments using two types of fingers: red and blue. Red fingers grasped objects horizontally, as depicted in Figure \ref{fig:instabil_conditions}(a). We adjusted the spacing between fingers to vary the grasping configuration for the same object. Furthermore, to examine the influence of object size, we conducted experiments on objects of different sizes while maintaining a consistent relative distance (spacing between fingers - object size).

\begin{figure}[t]
    \centering
    \subfloat[Blue fingers, change in offset, instability about $z$]
    {\includegraphics[width=0.9\columnwidth]{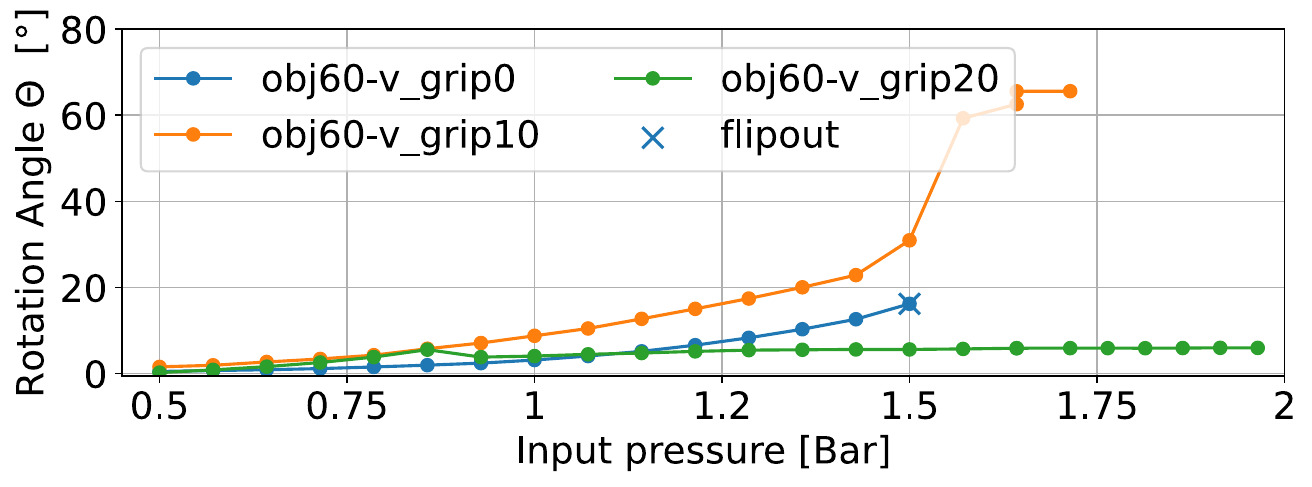} \label{fig:subim11}} \\
    \subfloat[Red fingers changing object diam, fixed offset, instability about $x$]
    {\includegraphics[width=0.9\columnwidth]{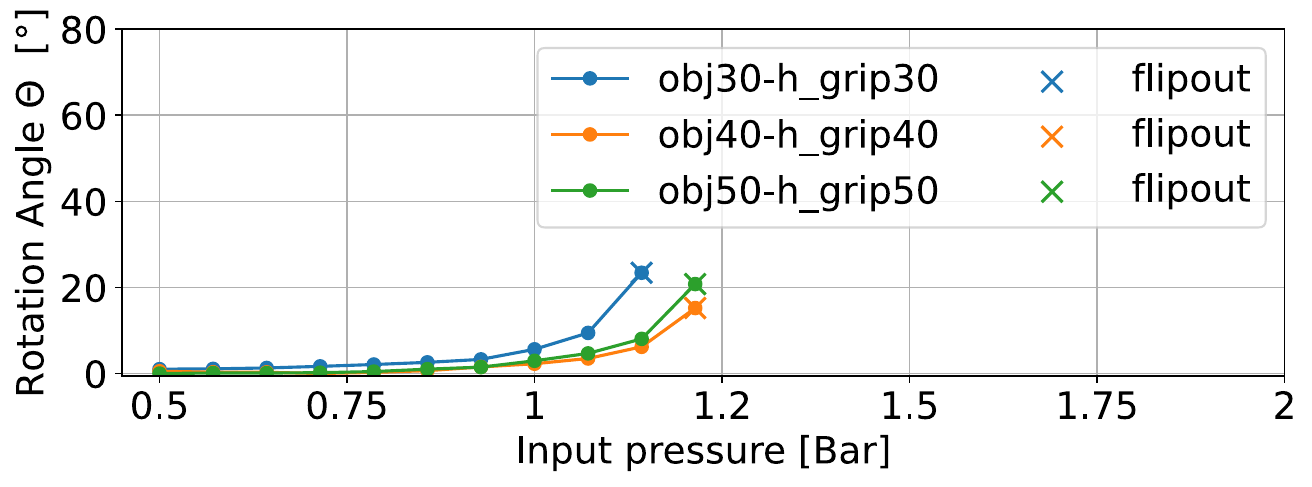} \label{fig:subim22}} \\
     \subfloat[Red finger, varying offset, instability about $x$]
    {\includegraphics[width=0.9\columnwidth]{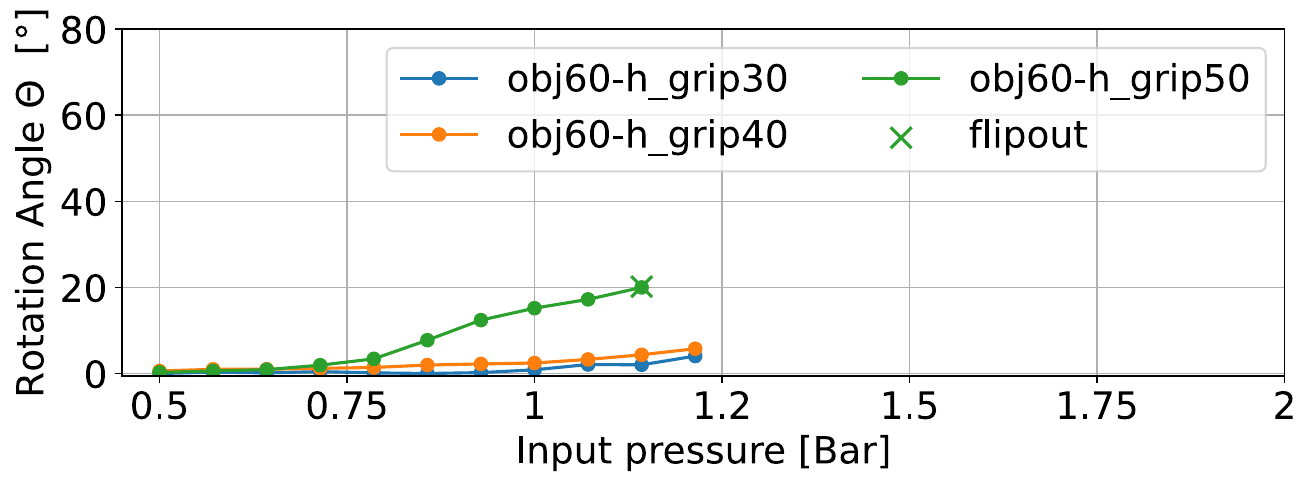} \label{fig:subim33}} \\
\caption{Rotation angle over pressure for the different grip conditions shown in Figure \ref{fig:instabil_conditions}. }
\label{fig:instabil_results}
\vspace{-10pt}
\end{figure}

In the case of blue fingers, objects were grasped vertically, as shown in Figure \ref{fig:instabil_conditions}(b). We conducted experiments with three different vertical offsets and measured angular displacement and instability based on pressure.
Figure \ref{fig:instabil_results} shows the measured rest angle of both fingers with the respective offsets. The progression curves clearly depict the anticipated trend from the model, with the flipout angle indicated at the corresponding point. Notably, the trend line varies according to the grip condition.

\begin{figure}[h]
    \centering
   \subfloat[Model instable force $f_p^i$ vs force measured at $5\deg$]{\includegraphics[width=0.8\columnwidth]{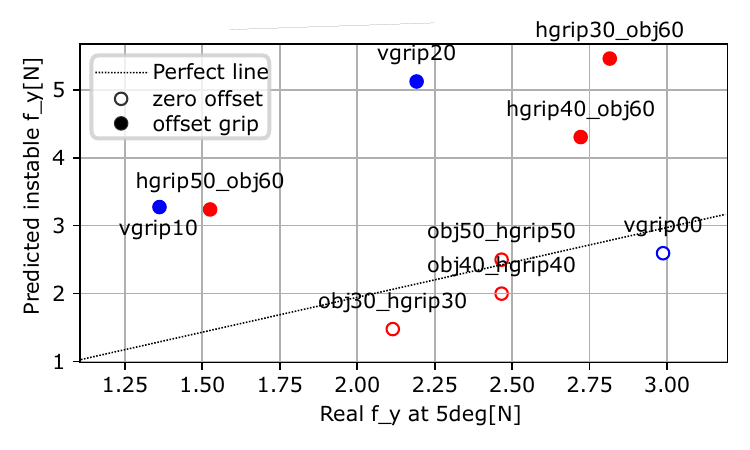}} \\
   \vspace{-5pt}
   \subfloat[Friction failure angle $\theta_f$ vs measured]{\includegraphics[width=0.8\columnwidth]{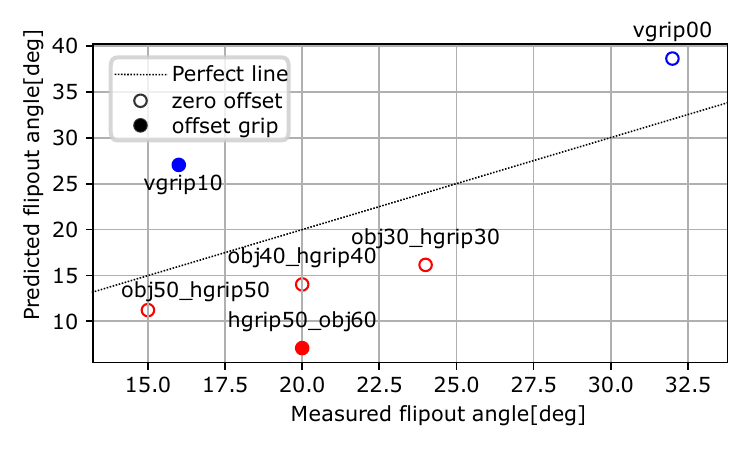}}\\
      \vspace{-5pt}
   \subfloat[Measured angle over model prediction rest angle]
    {\includegraphics[width=0.65\columnwidth]{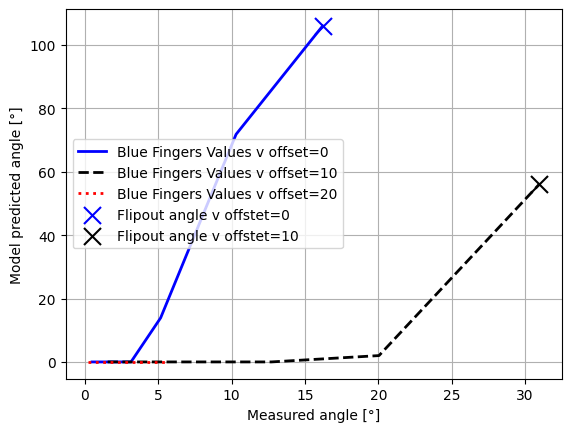}}\\
\caption{Comparing model predictions for: (a) model instability force $f_p^i$ compared with force when object reaches $5\deg$,  (b)  predicted flipout angle $\theta_f$ from \eqref{eq:friction_angle} vs measured flipout angle and (c) measured rest angle vs model rest angle.}
\label{fig:instabil_comparison}
\vspace{-15pt}
\end{figure}

Figure \ref{fig:instabil_comparison}(a) make comparison between predicted instability and the force when the object crosses $5\deg$ rotation in Figures \ref{fig:instabil_results}. While a trend can be seen, rotation often begins at a lower $f_p$ than predicted. Red and blue finger with zero offset grip has lower error between prediction and actual results than offset grasping. We attribute this mostly to asymmetric force application, where a non-zero transverse force from offset gripping induces rotation. In Figure \ref{fig:instabil_comparison}(b), the same comparison can be seen for the measured and predicted flipout angle.  Likewise, the zero offset grip matches more closely to the measured flipout angle than offset grip. 

Figure \ref{fig:instabil_comparison}(c) shows the measured rest angle over the model rest angle for the blue fingers. The stiffness of the red fingers did not meet the conditions to find a non-zero solution. If the model and reality match, the points should lie on the angle bisector. However, it can be seen that the model provides a more rapid increase and the later instability point. This is due to the fact that in the modeling a centered grip was assumed and no preload in the transverse direction was taken into account. In realistic grasps, there is typically imperfections in the grip mounting or differences in the fingers which lead to asymmetric force application and net forces in the transverse direction.

\subsection{Grip parameter optimization}
\begin{figure}[h]
    \centering
    \subfloat[Start, object in holder]{
        \includegraphics[trim={11cm 1cm 13cm 3cm}, clip, width=0.32\columnwidth]{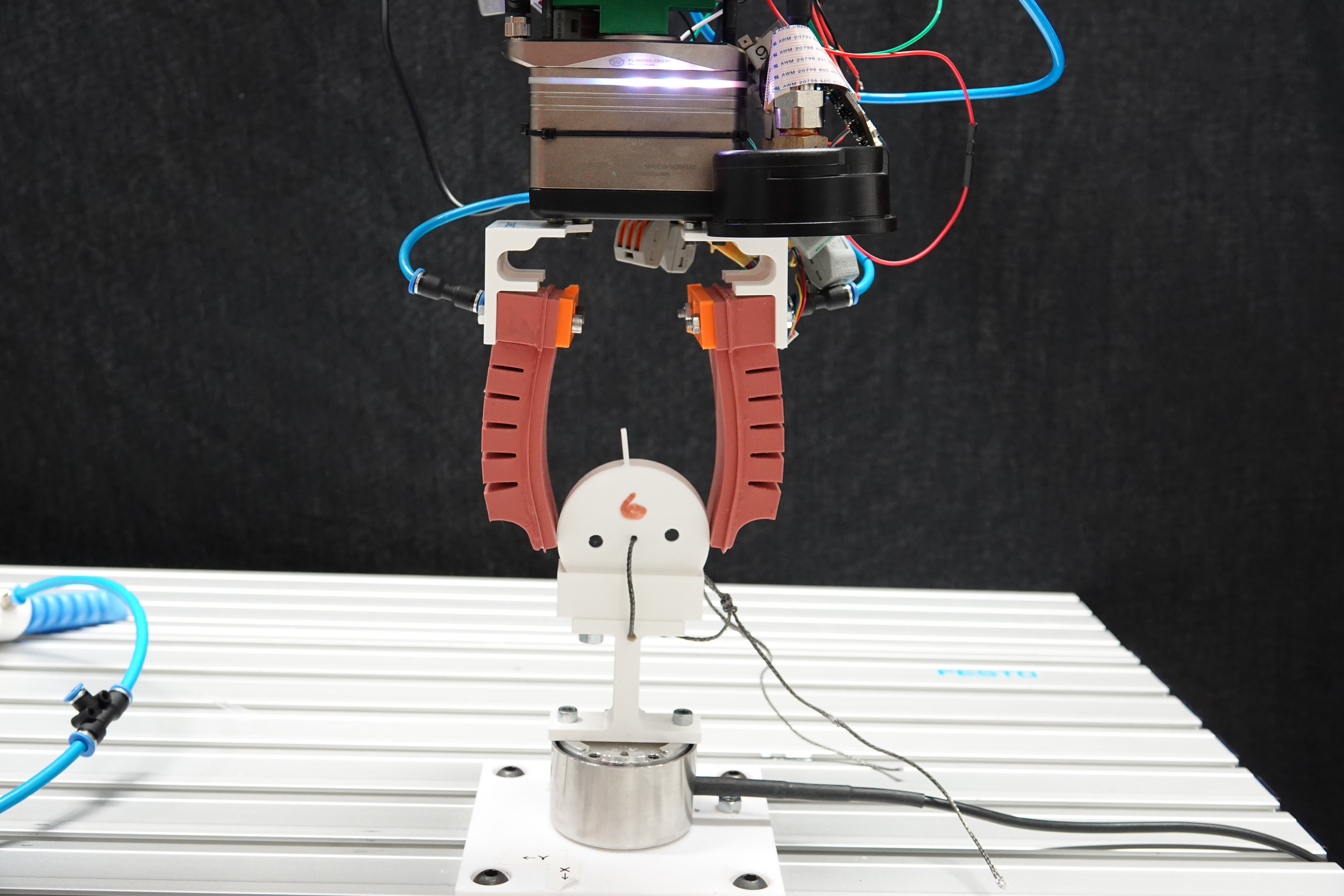}}
    \subfloat[Cable pulls as robot moves up]{        
        \includegraphics[trim={11cm 1cm 13cm 3cm}, clip, width=0.32\columnwidth]{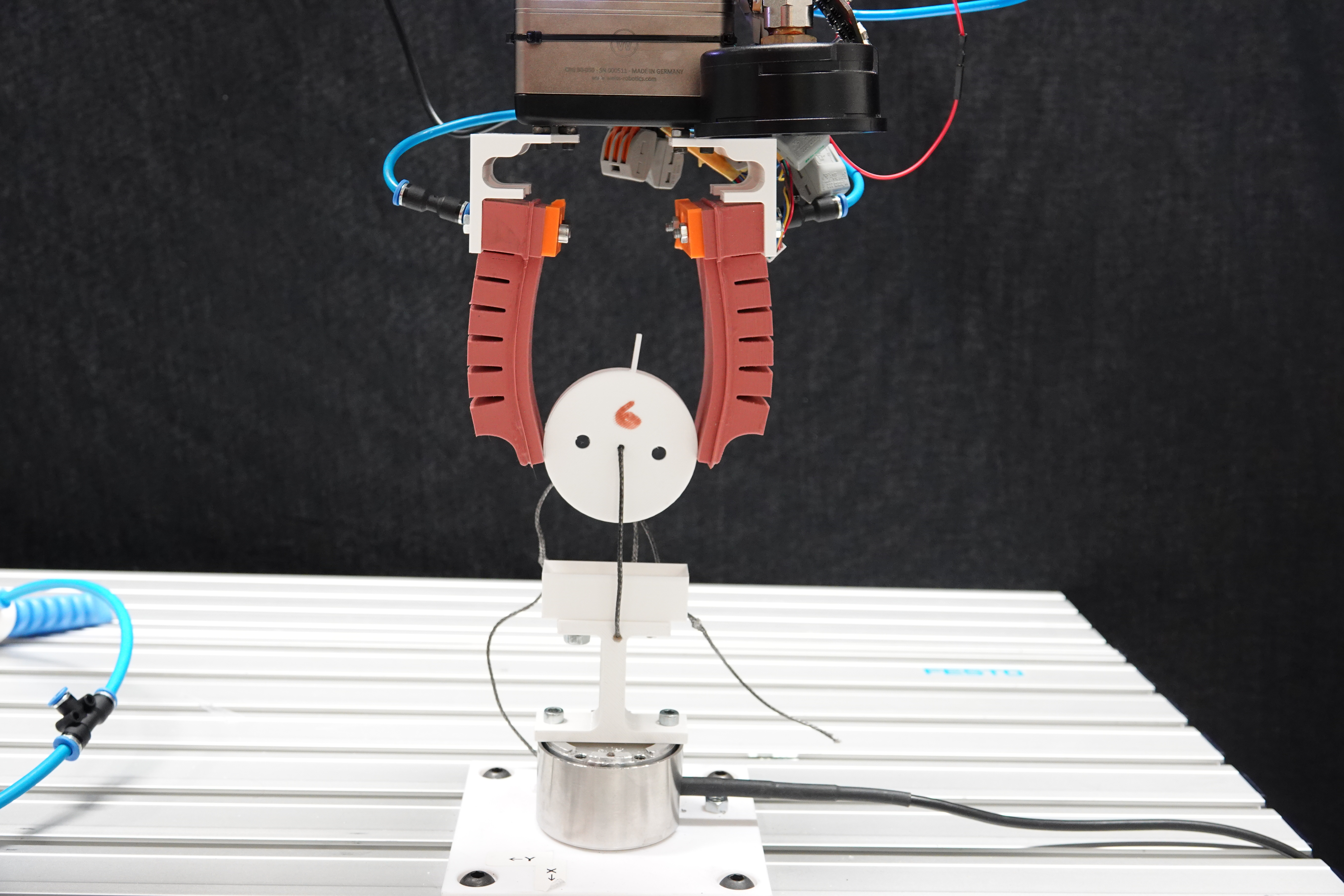}}
    \caption{Maximum pulling force setup, where the object starts in a holder and as the robot moves up, the cable pulls the object, transmitting forces to the F/T sensor while allowing instability \label{fig:max_grip}}
\end{figure}

\begin{figure}[h]
    \centering
            \includegraphics[ width=0.85\columnwidth]{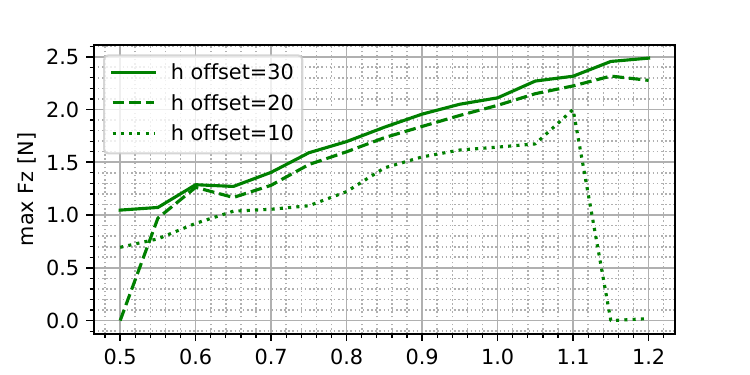}
    \caption{Maximum pulling force results, where the highest offset achieves the highest grasp force. \label{fig:max_grip_results}}
\end{figure}

Can the models be used to increase force capacity? This is investigated by experimentally finding the maximum pulling force in the vertical direction for the PneuNets, for a range of finger pressures and horizontal offsets. To measure force while allowing object instability, we modify the test objects to pass a string from the center, which is then fastened to the F/T sensor. In this way, if the grasp is unstable, the object will fly out and the resulting pull forces will be $0$. The grasp conditions are started as seen in Figure \ref{fig:max_grip}, and the robot moves vertically $30$mm, and the maximum force $f_z$ which occurs during this is measured. These experiments are repeated over a range of horizontal offsets and pressures, and the results can be seen in Figure \ref{fig:max_grip_results}. 

The most stable grasp, expected from Figure \ref{fig:stiff_measure}, is with horizontal offset of $30$mm which should be stable up to a grasp preload of $f_y=k_zr/2=4.8$N.  We find that this grasp also provides the highest pull force.

\section{Conclusion}
In conclusion, this study explores the rotational dynamics in soft fingers' precision grasping. By identifying grip failure modes—slip and dynamic rotational stability—the research lays the groundwork for optimizing grasp parameters. Analytical models developed in the study provide insights into factors influencing angular displacement, such as contact stiffness, normal force and radius of the object. The models are validated for the pressure at which rotation begins and when frictional failure occurs in various gripping conditions. In scenarios of zero offset grasping within two different finger systems, where transverse forces were absent, the model demonstrated relatively accurate predictions. Nevertheless, when applied for predictive purposes overall, limitations in accuracy were evident.

The model faces limitations due to strong assumptions, such as neglecting the viscoelastic effects, gravity, surface properties of objects, and ignoring any transverse forces $f_t$. Moreover, the measurement required for all contact variables poses a significant challenge, and the model is only validated for single, small contact patches with the objects, leaving questions on its ability to extend to multi-contact grasps.

{}

\clearpage

\bibliographystyle{IEEEtran}
\bibliography{lib_kevin,lib_valentyn,lib_hun}

\end{document}